\documentclass[sigconf]{acmart}

\usepackage{amsmath,amsfonts}
\usepackage{graphicx}
\usepackage{textcomp}
\usepackage{xcolor}
\usepackage[utf8]{inputenc}
\usepackage{amsfonts}
\usepackage{subcaption}
\usepackage{booktabs}

\usepackage{amsmath}

\usepackage{algorithm}
\usepackage{algorithmic}
\usepackage{clrs+,listings, url}
\usepackage{hyperref,xcolor}
\usepackage{graphicx, color}

\newcommand{\mA}{\mathbf{A}} 
\newcommand{\mZ}{\mathbf{Z}}
\newcommand{\mG}{\mathbf{G}}
\newcommand{\transpose}     {^{\mbox{\scriptsize \sf T}}}

\newcommand{\mW}{\mathbf{W}}
\newcommand{\mY}{\mathbf{Y}}
\newcommand{\mH}{\mathbf{H}}
\newcommand{\mT}{\mathbf{T}}

\newcommand{\dnnz}{\mathit{nnz}}

\copyrightyear{2024}
\acmYear{2024}
\setcopyright{rightsretained}
\acmConference[ICPP '24]{The 53rd International Conference on Parallel Processing}{August 12--15, 2024}{Gotland, Sweden}
\acmBooktitle{The 53rd International Conference on Parallel Processing (ICPP '24), August 12--15, 2024, Gotland, Sweden}
\acmPrice{}
\acmDOI{10.1145/3673038.3673152}
\acmISBN{979-8-4007-1793-2/24/08}
\begin{document}

\title{Sparsity-Aware Communication for Distributed Graph Neural Network Training}

\author{Ujjaini Mukhodopadhyay}
\affiliation{%
    \institution{University of California, Berkeley}
    \city{Berkeley}
    \state{CA}
    \postcode{94720}
    \country{USA}}
\email{ujjaini@berkeley.edu}
\author{Alok Tripathy}
\affiliation{%
    \institution{University of California, Berkeley}
    \city{Berkeley}
    \state{CA}
    \postcode{94720}
    \country{USA}}
\email{alokt@berkeley.edu}
\author{Oguz Selvitopi}
\affiliation{%
    \institution{Lawrence Berkeley Nat. Laboratory}
    \city{Berkeley}
    \state{CA}
    \postcode{94720}
    \country{USA}}
\email{roselvitopi@lbl.gov}
\author{Katherine Yelick}
\affiliation{%
    \institution{University of California, Berkeley}
    \city{Berkeley}
    \state{CA}
    \postcode{94720}
    \country{USA}}
\email{yelick@berkeley.edu}
\author{Ayd{\i}n Bulu\c{c}}
\affiliation{%
    \institution{Lawrence Berkeley Nat. Laboratory}
    \city{Berkeley}
    \state{CA}
    \postcode{94720}
    \country{USA}}
\email{abuluc@lbl.gov}

\begin{abstract}
Graph Neural Networks (GNNs) are a computationally efficient method to learn embeddings and classifications on graph data.
However, GNN training has low computational intensity, making communication costs the bottleneck for scalability. 
Sparse-matrix dense-matrix multiplication (SpMM) is the core computational operation in full-graph training of GNNs.
Previous work parallelizing this operation focused on sparsity-oblivious algorithms, where matrix elements are communicated regardless of the sparsity pattern.
This leads to a predictable communication pattern that can be overlapped with computation and enables the use of collective communication operations at the expense of wasting significant bandwidth by communicating unnecessary data.

We develop sparsity-aware algorithms that tackle the communication bottlenecks in GNN training with three novel approaches.
First, we communicate only the necessary matrix elements.
Second, we utilize a graph partitioning model to reorder the matrix and drastically reduce the amount of communicated elements.
Finally, we address the high load imbalance in communication with a tailored partitioning model, which minimizes both the total communication volume and the maximum sending volume.
We further couple these sparsity-exploiting approaches with a communication-avoiding approach (1.5D parallel SpMM) in which submatrices are replicated to  reduce communication.
We explore the tradeoffs of these combined optimizations and show up to $14\times$ improvement on $256$ GPUs and on some instances reducing communication to almost zero resulting in a communication-free parallel training relative to a popular GNN framework based on communication-oblivious SpMM.

\end{abstract}
\maketitle




\section{Introduction}
Graph neural networks (GNNs) have recently demonstrated success for many scientific and engineering problems, such as protein structure prediction in structural biology, traffic prediction in autonomous driving, and fraud detection~\cite{wu2020comprehensive}.
While GNNs are more compact than their dense counterparts, the graphs can be enormous, necessitating distributed training.
%
%
Sparse-matrix tall-skinny-dense-matrix multiplication (SpMM) has been identified as the bottleneck of GNN training, especially for the full-graph training case we consider in this paper.
We focus on full-batch GNN training over mini-batch training, as full-batch training GNNs offers many benefits over mini-batch training. 
Mini-batch training GNNs, where minibatches are batches of vertices, requires sampling from the $L$-hop neighborhood for each batch~\cite{hamiltonInductive2017,chen2018fastgcn,graphsaint}.
These sampling algorithms suffer from irregular memory accesses, lack of parallelism, and risk accuracy degradation, whereas full-batch training circumvents the need for sampling.
One way to parallelize full-graph GNN training on distributed-memory systems is to utilize a sparsity-oblivious approach where parts of the matrices that are communicated throughout the execution is independent of the graph structure.
Sparsity-oblivious approach has a number of benefits, such as straightforward generalization from dense matrix algorithms, ability to overlap communication with computation due to regular communication phases, and the ability to utilize  collective communication operations which often use less bandwidth than their point-to-point counterparts.

In this work, we show that all the benefits of the sparsity-oblivious approach listed above are far outweighed by a \emph{sparsity-aware} approach that takes advantage of the sparsity of the input graph for the purpose of reducing communication overheads.
Our sparsity-aware approach communicates only the necessary matrix blocks by avoiding communication of dense matrix elements corresponding to the empty columns in a local submatrix.
We extend this idea to all of GNN training steps.
Unfortunately, input graphs may not often come in a structure that maximizes this kind of format allowing the sparsity-aware approach to take advantage of. 


Graph and hypergraph partitioning has a long and celebrated history in scientific computing, which are documented in recent surveys~\cite{gpsurvey,ccatalyurek2022more}.
Perhaps one of the most canonical uses of graph/hypergraph partitioning is for sparse matrix-vector multiplication (SpMV) typically within an iterative sparse solver such as conjugate gradient.
However, the overhead of partitioning often takes many SpMV iterations to amortize, limiting the widespread use of partitioners in production codes.
Partitioning is also utilized to parallelize more heavy-weight kernels such as SpMM and sparse-matrix sparse-matrix multiplication~\cite{Ballard2015, Akbudak2018}, where it is often easier to amortize the overhead of the partitioning as these operations incur much more computation and communication than parallel SpMV.

In contrast to sparse iterative solvers, GNN training has significantly more work to do, easily making up for the cost of partitioning with the reduction in runtime.
This is because (i) the main workhorse SpMM is more expensive than SpMV, (ii) each epoch of training performs $2(L-1)$ SpMM operations where $L$ is the number of neural network layers, (iii) it takes hundreds of epochs for GNN training to converge to the desired accuracy, and (iv) the sparsity pattern of the matrix that represents the input graph does not change throughout training.
Hence, we can reorder the graph only once, which easily amortizes the cost of hundreds or thousands of parallel SpMM operations.

Graph and hypergraph partitioners by default optimize the total communication volume, while attempting to balance the computational load (e.g., by assigning the same number of nonzeros per processor in  the case of sparse matrix partitioning). 
However, most popular partitioners such as METIS~\cite{Karypis1998}, which are used by the most popular GNN training systems, do not provide a mechanism to balance the communication.
This is despite the fact that the bottleneck in distributed SpMM is the maximum communication volume between a pair processes. 
%
%
As the communication cost of the parallel SpMM is more volume-bound than it is latency-bound, due to large lengths of the feature vectors that need to be communicated, the load imbalance in communication can be severe and hurt scalability as we demonstrate in this paper.
%
To alleviate this issue, we rely on a recently-proposed partitioning method~\cite{acer2016improving} that aims to minimize the maximum amount of communicated data in addition to the total amount.
%

Our contributions in this paper are as follows:
\begin{itemize}
    \item For all steps of GCN (graph convolutional network) training, we present sparsity-aware algorithms that only communicate parts of dense matrices that result in nonzero output when multiplied with the sparse matrix. 
    \item We use graph partitioning to exploit the communication-reducing effect of sparsity-aware algorithms. We employ a specialized partitioner that is designed to minimize the maximum amount of communication between pairs of communicating processes, which is equivalent to minimizing communication load imbalance. 
    \item We demonstrate the generality of our approach by integrating the sparsity-awareness to both 1D and communication-avoiding 1.5D algorithms. 
    \item We demonstrate significant performance improvements for full-graph GNN training, compared to both the sparsity-oblivious approach as well as a sparsity-aware implementation that uses off-the-shelf partitioners that only consider the total volume.
\end{itemize}
%

The rest of this paper is organized as follows.
Section~\ref{sec:bg} introduces the necessary background and notation for parallel GNN training.
Section~\ref{sec:rw} surveys the studies on full-batch parallel training of GNNs.
Section~\ref{sec:spaware-gnn} presents our parallel 1D and 1.5D sparsity-aware algorithms for GNN training.
Section~\ref{sec:gp} examines graph partitioning models to distribute the matrices in the training and further reduce the communication overheads.
Finally, Section~\ref{sec:exp-setup} gives the experimental setup while Section~\ref{sec:results} presents the obtained results.

\section{Background}
\label{sec:bg}
\subsection{Graph Neural Networks}
GNNs take as input a graph $G = (V, E)$, where $V$ is the set of vertices and $E$ is the set of edges. This graph can represent any network structure found in the real world, such as protein-protein interaction, particle tracks, metagenomic read overlaps, social networks, transportation networks, etc. While GNNs can solve a wide variety of machine learning problems, we focus on \textit{node classification} without loss of generality. In this problem, each vertex takes an associated \textit{feature vector} as input, and a subset of vertices have an associated \textit{label}. The objective of the network is to classify unlabelled vertices in the graph using input features, graph connectivity, and vertex labels. To that end, the neural network maps vertices to low-dimensional embedding vectors such that similar vertices have similar embedding vectors. More concretely, the similarity of two vertices $u, v \in V$ with embedding vectors $z_u$ and $z_v$, respectively, is simply the dot-product $z_u\transpose z_v$. The feature vectors for each vertex are represented with a tall-skinny dense matrix $\mH \in \mathbb{R}^{n\times f}$. 

GNNs follow the \textit{message-passing} model, which consists of a \texttt{message} step and an \texttt{aggregate} step at each iteration of training~\cite{hamiltonInductive2017}. The \texttt{message} step creates a message for each edge in the graph. The \texttt{aggregate} step takes a vertex $v$ and combines the messages across all of that $v$'s incoming neighbors. The output is multiplied with a parameter weight matrix, and the result is an embedding vector $z_v$. Message-passing can be expressed in terms of sparse matrix multiplication $\mZ^l \gets \mA\transpose\mH^{l-1}\mW^l$. This formulation represents forward propagation in Graph Convolution Networks (GCNs), introduced by Kipf and Welling~\cite{KipfWelling2017}. In addition, forward propagation includes an activation function $\mH^l \gets \sigma(\mZ^l)$. After several layers of both steps, the network outputs an embedding vector per vertex, after which the network inputs vectors and labels into a loss function for backpropagation. Prior work has shown that the operations in GCN training, for both forward and backward propagation, are~\cite{tripathy2020reducing}
\begin{align*}
\mZ^{l} &\gets \mA\transpose\mH^{l-1}\mW^{l} \\
\mH^l &\gets \sigma(\mZ^l) \\
\mG^{l - 1} &\gets \mA \mG^{l}(\mW^{l})\transpose \odot \sigma '(\mZ^{l - 1}) \\
\mW^{l - 1} &\gets \mW^{l - 1} - \mY^{l - 1}.
\end{align*}
Here, the first two operations represent forward propagation in GCN training, while the last two compute the input and weight gradients, respectively. Int his work, we focus on optimizing GCNs, but all methods can be generalized to other types of GNNs.
\begin{table}[t]
\centering
\footnotesize
\caption{List of symbols and notations used by our algorithm} \label{symboltable}
\begin{tabular}{ |p{2cm}||p{5.5cm}|}
\hline
\multicolumn{2}{|c|}{Symbols and Notations} \\
\hline
Symbol & Description  \\
\hline
$\mA$ & Modified adjacency matrix of graph ($n \times n$)\\
$\mH^l$ & Embedding matrix in layer $l$ ($n \times f$)\\
$\mW^l$ & Weight matrix in layer $l$ ($f \times f$)\\
$\mA_i$ & $i$th row stripe of $\mA$ \\
$\mA_{ij}$ & The submatrix in the intersection of $i$th row stripe and $j$th column stripe of $\mA$ \\
$\sigma$ & Activation function\\
$f$ & Length of feature vector per vertex \\
$f_u$ & Feature vector for vertex $u$   \\
$L$ & Total layers in GNN \\
$P$ & Total number of processes \\
$\alpha$ & Latency \\
$\beta$ & Reciprocal bandwidth \\
\hline 
\end{tabular}
\end{table}

\section{Related Work}
\label{sec:rw}
\subsection{Graph Neural Network Systems}
Parallel training of GNNs is a recent and popular field of study, with many contributions from academia, government, and industry research labs. 
A recent ACM computings survey~\cite{abadal2021computing} focuses on the computational aspects of GNN training, and covers many systems and frameworks. 
In this work, we focus on full-graph training (as opposed to mini-batch training) and hence focus on works that support full-graph training. 
Dorylus~\cite{Dorylus} and DistGNN~\cite{md2021distgnn} are examples of distributed full-graph training on CPUs and ROC~\cite{jia2020improving}, CAGNET~\cite{tripathy2020reducing}, and BNS-GCN~\cite{wan2022bns} are examples of distributed full-graph training on GPUs. 
Dorylus, DistGNN, ROC, and BNS-GCN all use a vertex-centric perspective on GNN training, and use graph partitioning to minimize communication costs across devices and nodes. 
SpMM, which is the workhorse of full-batch GNN training, has been the focus of several works that investigate the performance of this operation under different settings.
CAGNET uses a matrix-centric perspective on GNN training, which distributes training by using distributed SpMM algorithms.
This approach has benefits, like provable communication bounds, but suffers from increased communciation volume by not factoring in the input graph's inherent sparsity.
Selvitopi et al.~\cite{Selvitopi2021} presented and investigated 1.5D and 2D sparsity-oblivious algorithms under bulk-synchronous and asynchronous communication scenarios in a setting with distributed-memory nodes equipped with GPUs.
Koanantakool et al.~\cite{spdmmm16} investigated communication costs of a number of 1.5D and 2D distributed SpMM algorithms and gave a recipe for which one to use according to different factors such as replication factor, relative sparsity, etc.
\subsection{Graph Partitioning}
Graph partitioning has been previously employed by several systems for reducing communication in GNN training~\cite{Zheng2020,jia2020improving,md2021distgnn,wan2022bns}. 
In a recent work~\cite{Merkel2023}, various vertex-based and edge-based graph partitioning models are investigated for distributed GNN training frameworks,  and it concluded that graph partitioning is crucial for optimizing parallel performance.
Most use METIS~\cite{Karypis1998} as the underlying partitioner. 
ROC uses its own partitioner with a dynamic programming approach.
%
All of these systems have advantages over matrix-centric approaches, like CAGNET, by leveraging the inherent sparsity in a graph to reduce communication.
However, the partitioning employed by each system reduces total volume of communicated data and overlooks the potential load imbalance in it, which can be a serious bottleneck in parallel training.
We investigate applying graph partitioners that minimize both total communication volume and the maximum communication volume between any pair of processes.
%

Graph/hypergraph partitioning in the context of SpMM has extensively been studied by Acer et al.~\cite{acer2016improving} in a distributed-memory setting, which proposed a general framework to encapsulate various communication cost metrics related to volume throughout the partitioning.
Among other partitioners that address multiple communication cost metrics and can be utilized in the parallelization of SpMM are Deveci et. al. and Slota et. al.~\cite{Deveci2015, Slota2014}.

In this work, we augment the matrix-centric approach by CAGNET by using distributed sparsity-aware SpMM algorithms.
This retains the benefit of having provable communication bounds while also reducing communication by exploiting the input graph's sparsity as present in other GNN systems.
In addition, we apply graph partitioners that minimize both the total communication and the maximum volume between any pair of processes, mitigating the load imbalance present in other GNN systems.
\section{Sparsity-Aware SpMM Algorithms}
\label{sec:spaware-gnn}
In this section, we present our 1D and 1.5D sparsity-aware distributed SpMM algorithms for full-batch GNN training. 
We focus on 1D and 1.5D algorithms as they outperformed other algorithms (e.g. 2D and 3D algorithms~\cite{Geijn1997,azad2016exploiting}) in CAGNET~\cite{tripathy2020reducing}.
However, these methods can be generalized to fit other algorihmic paradigms as well.
These algorithms are adapted and enhanced from their sparsity-oblivious counterparts~\cite{spdmmm16}. 
The memory costs of GNN training are dominated by the input graph $\mA$ and node activations $\mH^0, \ldots, \mH^{L-1}$, with a total cost of $O(\dnnz(\mA) + nfL)$. 
2D and 3D algorithms exist in the literature as well, however, they are shown to be less performant for full-batch GNN training.

Sparsity-aware algorithms improve their sparsity-oblivious counterparts by communicating less data and consequently improving runtime~\cite{ballard2013communication}.
In the sparsity-oblivious SpMM algorithms communication is performed on units of entire block rows of the dense matrix $\mH$.
This approach has regular communication patterns and can make use of collective operations.
However, the sparsity-oblivious algorithms unnecessarily communicate rows of $\mH$ that may not be needed in local SpMM computations due to likely empty columns in the local sparse matrix.
We introduce sparsity-aware algorithms that communicate only the necessary rows of $\mH$ in order to reduce communication overheads.
The two sparsity-aware algorithms presented in this section (Algorithms~\ref{alg:blockrow1dforward} and~\ref{alg:blockrow15dforward}) take as input
\begin{enumerate}
    \item $\mA \in \mathbb{R}^{n \times n}:$ sparse adjacency matrix,
    \item $\mH^{l-1} \in \mathbb{R}^{n \times f^{l-1}}:$ dense input activation matrix,
    \item $\mW \in \mathbb{R}^{f^{l-1} \times f^{l}}:$ dense training matrix,
\end{enumerate} 
and output $\mH^L: \mathbb{R}^{n \times f^l}$, dense output activation matrix.

\begin{figure}
	\centering
	\includegraphics[width=0.46\textwidth]{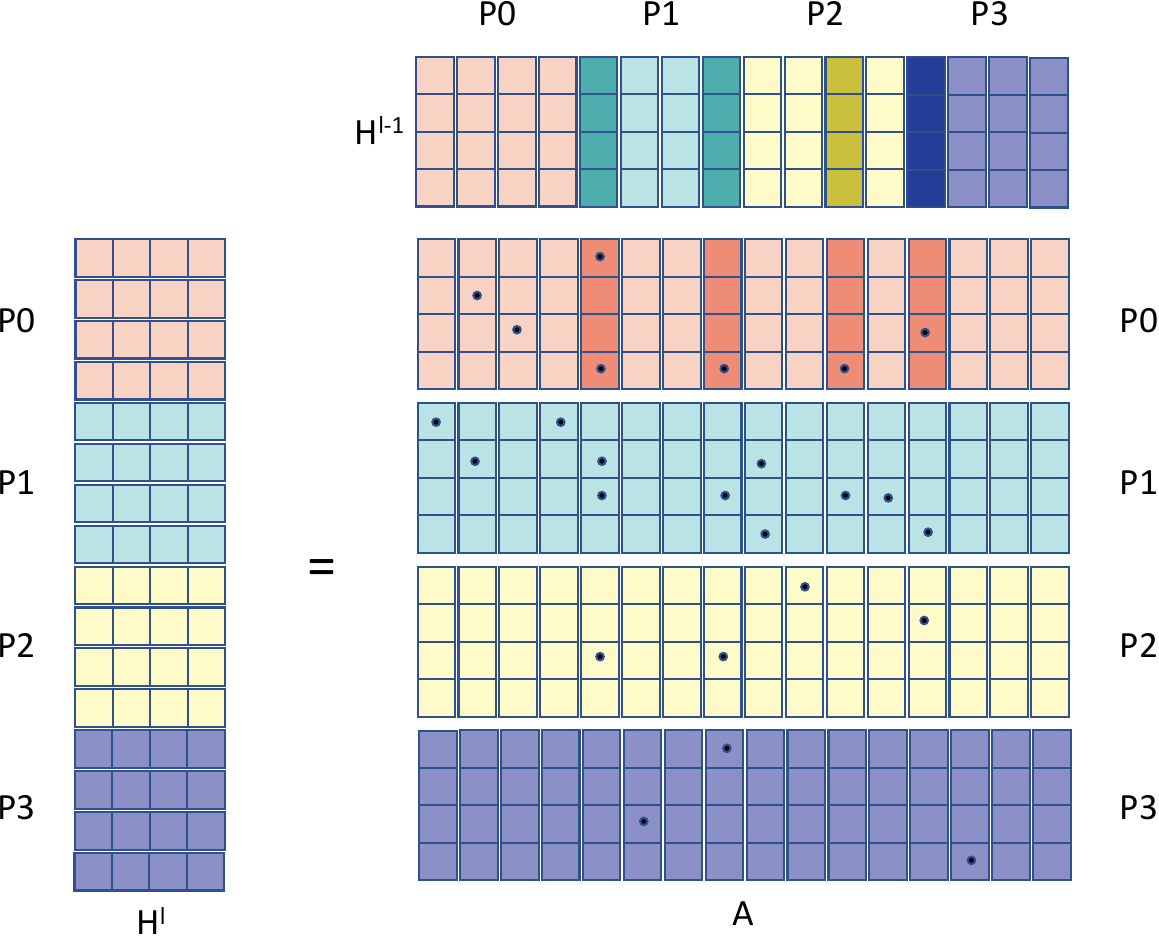}
	\caption{The partitioning of $\mA\transpose$ and $\mH$ in sparsity-aware 1D algorithm with $4$ processes. Boldly shaded columns in the first block row of $\mA\transpose$ and $\mH^{l-1}$ indicate the non-empty columns that require respective rows of $\mH$ needs to be received by P0.}
	\label{fig:sa-1d}
	\vspace{-1.5em}
\end{figure}

\subsection{Sparsity-Aware 1D Algorithm}
\label{sec:1d}
Our 1D algorithm assumes both $\mA\transpose$ and $\mH$ are distributed in block rows across processes, in which each process stores $n/P$ contiguous rows of $\mA\transpose$ and $\mH$. 
We use $\mA_i\transpose$ and $\mH_i$ to refer to the block rows of $\mA\transpose$ and $\mH$, respectively, and $\mA\transpose_{ij}$ to refer to the submatrix in the intersection of row $i$ and column $j$ on $P(i)$:

\begin{equation}
\mA\transpose = \left( 
\begin{array}{c}
\mA\transpose_{1} \\
\vdots \\
\mA\transpose_{p}
\end{array} 
\right)
= \left( 
\begin{array}{c c c}
\mA\transpose_{11} & \ldots  & \mA\transpose_{1 p} \\
\vdots  & \ddots  & \vdots  \\
\mA\transpose_{p 1} & \ldots   & \mA\transpose_{p p} 
\end{array} 
\right), 
\mH = \left( 
\begin{array}{c}
\mH_{1} \\
\vdots \\
\mH_{p}
\end{array} 
\right).
\label{eqn:1dpartitioning}
\end{equation}

Our sparsity-aware algorithms first has each process $P(i)$ locally compute the nonzero column indices in its row block in order to find out which rows of $\mH$ it needs.
For any $i,j$ pair, let $NnzCols(i, j)$ be a vector of nonzero column indices in $\mA\transpose_{ij}$. 
These indices specify the rows of $\mH$ needed to compute $\mA\transpose_{ij}\mH_j$, hence the rows that need to be received by $P(i)$.
Figure~\ref{fig:sa-1d} illustrates the distribution of the input matrices $\mA\transpose$ and $\mH$ among four processes in sparsity-aware 1D algorithm.
Boldly shaded columns in the first block row of $\mA\transpose$ and $\mH^{l-1}$ indicate the non-empty columns that require respective rows of $\mH$ needs to be received by P0.

Let $\mZ^l$ be the intermediate product $\mA\transpose\mH^{l-1}$. 
In our sparsity-aware 1D algorithm, $\mZ^l_i$ for process $P(i)$ is given by
$$ \mZ^l_i = \mZ^l_i + \mA\transpose_i \, \mH = \mZ^l_i + \sum_{j=1}^{p} \mA\transpose_{ij}  \, \mH_j. $$ 
Note that $P(i)$ stores $\mA\transpose_i$ locally, but not $\mH_j$ for $i\neq j$. 
In the sparsity-oblivious algorithm, each process $P(j)$ would broadcast its entire block row $\mH_j$ to all other processes. 
Our sparsity-aware 1D algorithm ensures that each process $P(j)$ only sends the necessary rows of $\mH$ to each other process using the computed nonzero column indices. 
Algorithm~\ref{alg:blockrow1dforward} describes how to compute $\mZ^l$.
Next, we analyze the communication requirements of the operations in GNN training.

\paragraph{Equation $\mZ^l = \mA\transpose \mH^{l - 1}\mW^l$.}
In Algorithm~\ref{alg:blockrow1dforward}, the only communication is an all-to-allv call that exchanges rows of $\mH$ (recall that $\mW^l$ is fully-replicated, so no communication is necessary).
Each process $P(i)$ needs to receive data from $P-1$ other processes, and the time taken by this operation can be upper-bounded by the maximum of size of $NnzCols(i, j)$ times $f$, for any $i,j$, which we denote with $cut_P(G)$ . 
%
%
This results in the following per-process communication cost with the $\alpha-\beta$ model:
$$T_{comm} = \alpha (P-1) + (P - 1) \; cut_P(G) \; f \;\beta$$.

\paragraph{Equation $\mH^l = \sigma(\mZ^l)$.} This operation is communication-free as $\mH^l$ is partitioned by rows.
%

\begin{algorithm}[tb]
\begin{algorithmic}
\FOR{all processes $P(i)$ \InParallel}
\STATE $\mT \gets [\mH[\textit{NnzCols}(i, 0)]; \ldots; \mH[\textit{NnzCols}(i, P-1)]$
\STATE $\textit{AllToAllv}(\mT, P(:))$
\FOR{$k = 0$ to $p - 1$}
    \STATE Initialize $\hat{\mH}^{l-1}$ 
    \STATE $\hat{\mH}^{l-1}[\textit{NnzCols}(k, i)] \gets \mT[k]$ 
    \STATE $\mZ^l \gets \mZ^l + \textit{SpMM}(\mA\transpose_{ij},\hat{\mH}^{l-1})$
\ENDFOR
\STATE $\mH^l \gets \textit{GEMM}(\mZ^l, \mW)$
\ENDFOR
\end{algorithmic}
\caption{Sparsity-Aware 1D algorithm for GNN forward propagation. 
} \label{alg:blockrow1dforward}
\end{algorithm}

\paragraph{Equation $\mG^{l-1} = \mA\mG^l(\mW^l)\transpose \odot \sigma'(\mZ^{l-1})$.} 
The communication in this step is identical to the communication in forward propagation.
For undirected graphs we have $\mA = \mA\transpose$ and no communication is needed for transpose.
For directed graphs, we store both $\mA$ and $\mA\transpose$.
In addition, $\mA$, $\mG^l$, and $\mZ^{l-1}$ are all partitioned into block rows.
Since $\mA$ is sparse and $\mG^l$ is dense, we use our sparsity-aware SpMM implementation to compute $\mA\mG^l$. 
The communication pattern is identical to that of in Algorithm~\ref{alg:blockrow1dforward}.

\paragraph{Equation $\mY^{l-1}= (\mH^{l - 1})\transpose \mA \mG^l$.}
Communication in this step is a small 1D outer product. 
Locally multiplying $\mH^{l-1}\mA\mG^l$ yields a single matrix per process of size $f\times f$ that must be reduced across processes. 
We treat this communication cost as a lower-order term.
\paragraph{Total Communication}
Ignoring lower order terms, such as reducing $f\times f$ matrices, the total communication cost for our sparsity-aware 1D algorithm is
$$T_{comm} = 2L(\alpha (P-1) + (P - 1) \; cut_P(G) \; f \;\beta)$$.

\begin{figure}
	\centering
	\includegraphics[width=0.49\textwidth]{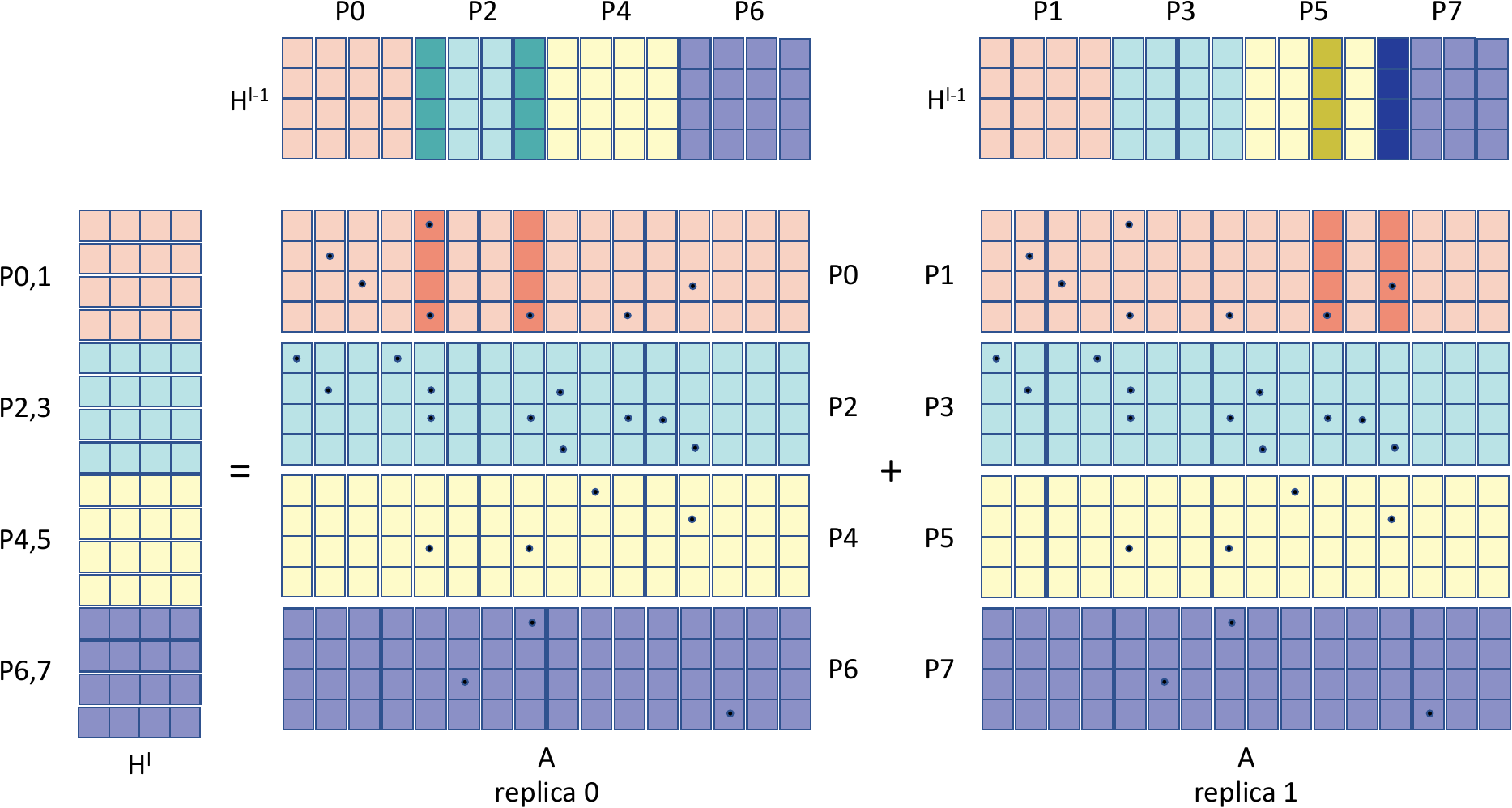}
	\caption{The partitioning of $\mA\transpose$ and $\mH$ in sparsity-aware 1.5D algorithm among eight processes with a replication factor of 2 ($c=2$) in sparsity-aware 1.5D algorithm. Boldly shaded columns in the first block row of $\mA\transpose$ and $\mH^{l-1}$ indicate the non-empty columns that require respective rows of $\mH$, which will be received by P0 and P1.}
	\label{fig:sa-15d}
\end{figure}

\subsection{Sparsity-Aware 1.5D Algorithm}
In 1.5D algorithms, processes are organized in a $P/c\times c$ process grid~\cite{spdmmm16,tripathy2020reducing}. 
In this regime, both $\mA\transpose$ and $\mH$ are partitioned into $p/c$ block rows, and each block row is replicated on $c$ processes. 
This memory replication oftentimes makes 1.5D algorithms have lower communication costs over 1D algorithms.
Specifically, each process in process row $P(i,:)$ stores $\mA\transpose_{i}$ and $\mH_i$:
\begin{equation}
\mA\transpose = \left( 
\begin{array}{c}
\mA\transpose_{1} \\
\vdots \\
\mA\transpose_{p/c}
\end{array} 
\right)
\mH = \left( 
\begin{array}{c}
\mH_{1} \\
\vdots \\
\mH_{p/c}
\end{array} 
\right).
\label{eqn:15dpartitioning}
\end{equation}
Each submatrix $\mA\transpose_{i}$ is further partitioned in $p/c$ block columns.
Let $\mT$ be the intermediate product $\mA\transpose\mH^{l - 1}$. 
Each process row $P(i, :)$ computes
$$ \mT_i = \mT_i + \mA\transpose_i \, \mH = \mT_i + \sum_{j=1}^{p/c} \mA\transpose_{ij}  \, \mH_j.$$
However, each process only computes a partial sum in this summation.
Of the $p/c$ terms, each process in $P(i,:)$ computes $q=p/c^2$ distinct terms in parallel.
These partial results are then summed across all $c$ processes in $P(i,:)$ with an all-reduce operation.
The result is the final $\mT_i$ matrix, which is replicated on each process in $P(i,:)$.
The computation performed by $P(i, j)$ is
\begin{equation}
\mT_i = \mT_i + \mA\transpose_i \, \mH = \mT_i + \sum_{k=jq}^{(j + 1)q} \mA\transpose_{ik}  \, \mH_k .
\label{eqn:15dsum}
\end{equation}

As in the 1D algorithm, $P(i, j)$ stores $\mA\transpose_{ik}$ locally, but accessing $\mH_k$ for all values of $k \neq i$ requires communication.
We refer to the blocks $\mA_{ik}\transpose$ where $i\neq k$ as the \textit{off-diagonal} blocks of $\mA\transpose$, and the nonzero columns in these blocks yield communication.
Figure~\ref{fig:sa-15d} illustrates the distribution of $\mA\transpose$ and $\mH$ among eight processes with a replication factor of 2 ($c=2$) in sparsity-aware 1.5D algorithm.
Boldly shaded columns in the first block row of $\mA\transpose$ and $\mH^{l-1}$ indicate the non-empty columns that require respective rows of $\mH$ needs to be received by P0 and P1.

The sparsity-oblivious algorithm for 1.5D partitioning communicates entire block rows of $\mH$.
Our sparsity-aware algorithm, on the other hand, communicates only the rows of $\mH$ needed for local SpMM computation, which are given by the nonzero column indices in the local matrices, and with an additional space and latency cost of communicating the necessary row indices.
Algorithm~\ref{alg:blockrow15dforward} describes how to compute $\mZ^l$.
Next, we analyze the communication requirements of these operations in GNN training.

\begin{algorithm}[t]
\begin{algorithmic}[1]
\FOR{all processes $P(i,j)$ \InParallel} 
\STATE $s = p / c^2$  \COMMENT{(number of stages)}
\FOR{$k = 0$ to $s - 1$}
    \STATE $q = j \, s + k$
    \IF{$P(i,j) = P(q,j)$}
        \FOR{$l = 0$ to $p/c$}
            \STATE $srows \gets \textit{NnzCols}(l,j)$
            \STATE $\textit{ISend}(\mH^{l-1}[srows, :], P(l,j))$
        \ENDFOR
    \ENDIF
    \STATE $rrows \gets \textit{NnzCols}(i,q)$
    \STATE $\textit{Recv}(\hat{\mH}^{l-1}[rrows,:], P(q,j))$
    \STATE $\hat{\mZ}^l \gets \hat{\mZ}^l + \textit{SpMM}(\mA\transpose_{iq},\hat{\mH}^{l-1})$
\ENDFOR
\STATE $\mZ^l \gets \textit{AllReduce}(\hat{\mZ}, +, P(i, :))$
\STATE $\mH^l \gets \textit{GEMM}(\mZ^l, \mW)$
\ENDFOR
\end{algorithmic}
\caption{Sparsity-Aware 1.5D algorithm for GNN forward propagation.
} \label{alg:blockrow15dforward}
\end{algorithm}

\paragraph{Equation $\mZ^l = \mA\transpose \mH^{l - 1}\mW^l$.}
\label{sec:15dfwd}
Our sparsity-aware 1.5D algorithm is presented in Algorithm~\ref{alg:blockrow15dforward}. 
Each iteration $q$ of the outer loop for process $P(i, j)$ receives the rows of $\mH$ at row indices $NnzCols(i, q)$, followed by a local SpMM.
The number of rows received is upper bounded by $cut_P(G)$.
Finally, the all-reduce has each process row reduce matrices of size $n/(p/c)\times f$. 
This yields an overall communication cost of
$$T_{comm} = \alpha \Big(\frac{P}{c^2}\log\frac{P}{c^2}\Big) + \frac{P}{c^2} \; cut_P(G) \; f \; \beta.$$
In practice, $cut_P(G)$ will scale roughly by $P/c$ since the graph partitioner partitions $\mA$ into $P/c$ partitions. 
Having $P/c$ in the denominator of $cut_P(G)$ yields a bandwidth term that scales by $c$.

\paragraph{Equation $\mH^l = \sigma(\mZ^l)$.}
This operation is communication-free as $\mH^l$ is partitioned by rows.

\paragraph{Equation $\mG^{l-1} = \mA\mG^l(\mW^l)\transpose \odot \sigma'(\mZ^{l-1})$.}
As in the 1D algorithm, no communication is required to transpose $\mA\transpose$ to get $\mA$, since we either assume a symmetric matrix or explicitly store the transpose if that is not the case.
Matrices $\mA$, $\mG^l$, and $\mZ^{l-1}$ are all partitioned in block rows, and $\mA$ is sparse while $\mG^l$ is dense. 
Thus, the communication operations of $\mA\mG^{l-1}$ are the same with the 1.5D algorithm described above.

\paragraph{Equation $\mY^{l-1}= (\mH^{l - 1})\transpose \mA \mG^l$.}
As in the 1D algorithm, communication in this step is a small outer product and locally multiplying $(\mH^{l-1})\transpose\mA\mG^l$ yields a single matrix per process of size $f\times f$ that must be reduced across processes, which we treat again as a lower-order term in communication.

\paragraph{Total Communication}
Ignoring lower-order terms, the total communication for our sparsity-aware 1.5D algorithm is 
$$T_{comm} = 2L\Big(\alpha \Big(\frac{P}{c^2}\log\frac{P}{c^2}\Big) + \frac{P}{c^2} \; cut_P(G) \; f \; \beta)\Big).$$

\section{Graph partitioning}
\label{sec:gp}
Graph partitioning and distributing both the $\mA$ and $\mH$ input matrices are critical parts of GNN training in order to reduce communication. Here, we discuss multiple approaches to graph partitioning, and outline why a partitioner must reduce both the total communication volume and maximum volume between two processes.

One approach for distributing both the sparse and dense matrices in both sparsity-oblivious and sparsity-aware GNN training is by randomly permuting the adjacency matrix $\mA$ and following a simple 1D block distribution where each block has roughly the same number of rows.
Although this approach can achieve somewhat good computational load balance, it has two main shortcomings that can hinder scalability.
First, it does not attempt to minimize the amount of communication during the training.
The amount of communication is dictated by the number of nonzeros in off-diagonal blocks of $\mA\transpose$, and simply cutting $\mA\transpose$ into block rows does not reduce this number.
This is valid even for the sparsity-aware training despite the fact that it selectively communicates only the rows of $\mH$ that are needed by a process.
A random permutation that is applied prior to training for achieving good computational load balance may exacerbate this issue as it may cause many nonzero column segments in off-diagonal blocks of $\mA$ -- which determine which rows of $\mH$ to communicate.
Another shortcoming is that an even distribution of rows of $\mA$ may not always yield good computational load balance if the number of nozeros in these rows are not even, which is usually the case in real-world graphs.
Both of these issues can be remedied by distributing the matrices with a graph partitioner.

\begin{table}[!tbp]
  \caption{Average and maximum amount of data communicated in a single SpMM where the sparse matrix is distributed with METIS graph partitioner (instance: Amazon, $f=300$).}
  \begin{center}
    \scalebox{1.0} {
      \begin{tabular}{c r r r}
        \toprule
         & \multicolumn{2}{c}{data size (MB)} &  \\
          \cmidrule(lr){2-3} 
         $p$ & average & max & load imbalance \% \\
         \midrule                                                              
        16  & 199.6 & 333.5 &  67.1\% \\
        32  & 132.9 & 241.6 &  81.8\% \\   
        64  &  83.9 & 164.0 &  95.4\% \\
        128 &  52.5 & 117.3 & 123.3\% \\
        256 &  32.6 &  86.4 & 164.9\% \\        
        \bottomrule
        \end{tabular}
    }
  \end{center}
  \label{tb:comm-imb-amazon-metis}
\end{table}

In sparsity-aware GNN training, $P(i)$ must receive rows of $\mH_j$ corresponding to the nonzero column segments in its off-diagonal blocks $\mA\transpose_{ij}$, where $i \neq j$.
Compared to sparsity-oblivious training, the sparsity-aware training aims to avoid receiving the entire $\mH_j$ by not communicating the rows of $\mH_j$ corresponding to the zero column segments.
However, if there are not many zero column segments in off-diagonal blocks, as is expected to happen if the graph is randomly permuted to get good computational load balance, the sparsity-aware training may not yield a big reduction in communication time.
Partitioning the adjacency matrix with a graph partitioner prior to training among $p$ processes helps reduce the number of nonzero column segments in off-diagonal blocks, in addition to achieving computational load balance.
Graph partitioning is commonly used to parallelize sparse iterative solvers, usually focusing on distributing the computations related to SpMV .
The partitioning models for SpMV can easily be extended to SpMM, which is the common operation in full-batch GNN training in this work. However, compared to SpMV, distributing the adjacency matrix for SpMM should be done more carefully as any imbalance in nonzeros assigned to each process amplifies the communication cost by at most $f$.

Among the two factors mentioned above, the first can easily be addressed by enforcing a stricter load balance constraint in partitioning.
The second factor of communication load imbalance is more difficult to address as most partitioners usually only aim at reducing the total edgecut in partitioning, which corresponds to reducing total amount of transferred data.
The problem of high load imbalance in communication can be severe as the overall communication time is determined by the bottleneck process, i.e., the process that communicates the maximum amount of data.
Table~\ref{tb:comm-imb-amazon-metis} presents various statistics regarding communication in a single SpMM obtained by using the partitioner METIS~\cite{Karypis1998} on Amazon data in GNN training for $p\in\{16,32,64,128,256\}$.
The large sizes of messages in megabytes coupled with communication load imbalance which can be as high as 165\% (i.e., the bottleneck process sending 2.7x the amount of data of an average process) makes it imperative to address this issue in order not to make communication a bottleneck.


%

%
To alleviate this issue, we use Graph-VB (GVB), a partitioner that can handle multiple communication cost metrics related to volume~\cite{acer2016improving}.
This partitioner can simultaneously handle metrics such as total volume of communicated data, maximum send volume, maximum receive volume, etc. 
In our work we rely on this partitioner to optimize the total and maximum send volume metrics.


\section{Experimental Setup}
\label{sec:exp-setup}
\subsection{System Details}
All our experiments are run on the Perlmutter supercomputer system at NERSC, where each node is equipped with 4 NVIDIA A100 GPUs with 40GB HBM memory on each node. 
We run exactly one process per GPU in our experiments.
In addition, there are 4 NVLink links between each pair of GPUs within a node each with a bandwidth of 25GB/s.
Each GPU is connected to an AMD EPYC 7793 CPU with a PCIe 4.0 bus.
The nodes possess 4 HPE Slingshot 11 NICs also using a PCIe 4.0 bus.
Each NIC supports a 25GB/s bandwidth.
%

\subsection{Implementation Details}
We use PyTorch 1.11 and PyTorch's \texttt{torch.distributed} package with a backend in NCCL 2.11.4 for distributed communication and CUDA 11.7. 
We start by partitioning adjacency matrix into block rows, and rearranging the rows of $\mH$ to match the new vertex ids post partitioning.
These block rows have variable size, depending on the output partition sizes.
%
%
This rearranging is done with several all-to-all calls as a preprocessing step, and the preprocessing time is significantly smaller than the training time.
Thus, we do not include it as part of our training times.

We use a 3-layer Graph Convolution Network (GCN) architecture, as introduced in Kipf and Welling ~\cite{KipfWelling2017}, with 16 hidden units and 100 epochs for training.
%
%
%
As mentioned in Section~\ref{sec:1d}, the 1D algorithm uses all-to-all communication.
This collective communication operation is optimized in NCCL with separate nonblocking point-to-point send and recv calls between each pair or processes.
Specifically, the all-to-all uses NCCL's \texttt{ncclGroupStart()} and \texttt{ncclGroupEnd()} functions, which surround pairwise \texttt{ncclSend()} and \texttt{ncclRecv()} calls and is used within the \texttt{torch.distributed} API ~\cite{ncclrepo}. 
Our 1.5D algorithm utilizes non-blocking sends (i.e., \texttt{Isend}s) and blocking recvs (i.e., \texttt{Recv}s). 
We use PyTorch's \texttt{batch\_isend\_irecv} API, which uses NCCL's \texttt{ncclGroupStart()} and \texttt{ncclGroupEnd()} functions to advantage of the same grouping behavior as the all-to-all calls.
For any local SpMM call, we use cuSPARSE's \texttt{csrmm2} function. 
%

We also compared the accuracy of the proposed sparsity-aware implementations of the 1D and 1.5D algorithms and the preceding sparsity-oblivious implementations, and we observed no change in accuracy apart from floating-point rounding errors.
This is expected as we have not changed the underlying multiplication operations.
Thus, in the following sections, we only focus on the performance benefits. 
In addition, we use the sparsity-oblivious implementations from Tripathy et. al. when comparing our approach to sparsity-oblivious algorithms~\cite{tripathy2020reducing}.

\subsection{Datasets}
We ran experiments for the 1D and 1.5D algorithms on the Reddit, Amazon, Protein, and Papers datasets.
Table~\ref{tab:data} presents basic properties of these graphs.
Each vertex of the Reddit graph represents a post and an edge exists between two vertices if the same user commented on both posts ~\cite{hamiltonInductive2017}.
This is our smallest and densest dataset.
The vertices of the Amazon graph encode different products, and an edge exists if there exists a buyer that purchased both products ~\cite{jia2020improving}.
This is our sparsest dataset.
The vertices of the Protein graph represents proteins, and there exists an edge between two vertices if the respective proteins exhibit a certain degree of similarity.
%
%
For the Protein graph, we used an induced subgraph consisting of 1/8 of the vertices of the larger original graph in HipMCL~\cite{hipmcl}.
Finally, the Papers dataset is a citation network where each vertex is an arXiv paper and each edge connects two papers if one paper cites the other~\cite{hu2020ogb}.
This is the largest graph we evaluate.
All three graphs are symmetric, thus we assume $\mA = \mA\transpose$ and only store the adjacency matrix once.

For the Reddit and Papers dataset, we used the original features and labels as in ~\cite{hamiltonInductive2017,hu2020ogb}. For Amazon and Protein datasets, we chose an arbitrary number of features and labels for each dataset, and use the adjacency matrix to encode the relationship between vertices. 

\begin{table}[t]
\centering
\caption{Datasets used in our experiments}
\label{tab:data}
\scalebox{0.90}{
\begin{tabular}{ |l|r|r|r|r|}
\hline
Graph & Vertices & Edges & Features & Labels  \\
\hline
Reddit & 232,965 & 114,848,857 & 602 & 41\\
Amazon & 14,249,639 & 230,788,269 & 300 & 24 \\
Protein & 8,745,542 & 2,116,240,124 & 300 & 24 \\
Papers & 111,059,956 & 3,231,371,744 & 128 & 172 \\
\hline
\end{tabular}
}
\end{table}

\begin{figure*}[!t]
    \centering
    \includegraphics[scale=0.38]{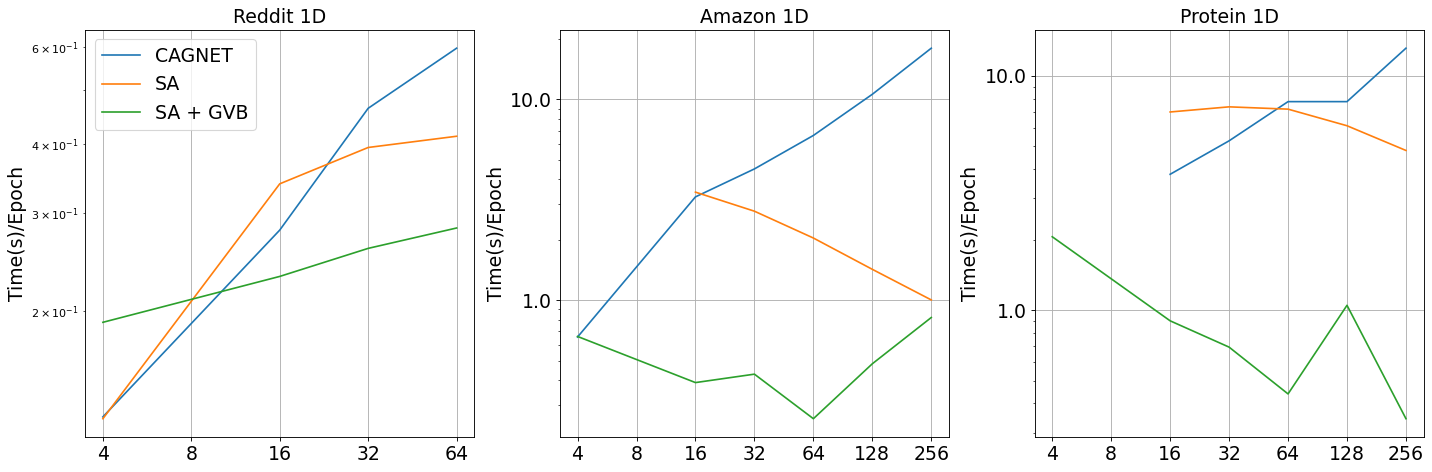}
    \caption{1D performance results for sparsity-oblivious, sparsity-aware, and sparsity-aware + GVB graph partitioning implementations. Note that these are log-log plots of the number of GPUs versus the time for a single epoch. For Reddit, we use $p=4, 16, 32, 64$. For Amazon and Protein datasets, we also use $p=128$ and $256$. Missing data in the line segments on Amazon for $p=4$ and on Protein for $p=4$ means that this trial of the experiment ran out of memory.}
    \label{fig:1dline}
\end{figure*}

\subsubsection{Graph Partitioning}
When we use a partitioner, the sparse matrix is permuted and distributed according to the partitioning output in which the rows and columns of the sparse matrix are ordered according to the relabeled vertex indices provided by the partitioner.
We use a symmetric permutation of the sparse matrix.
For these experiments, we use the partitioner Graph-VB~\cite{acer2016improving}, which optimizes both total and and maximum communication volume.
We also include results with METIS for comparison with a graph partitioner that only minimizes total communication.
We only run partitioning once once since the pattern of the sparse adjacency matrix does not change throughout GNN training.
%

\section{Results}
\label{sec:results}
\subsection{Performance of 1D Algorithm}
\begin{figure*}[!t]
    \centering
    \includegraphics[scale=0.45]{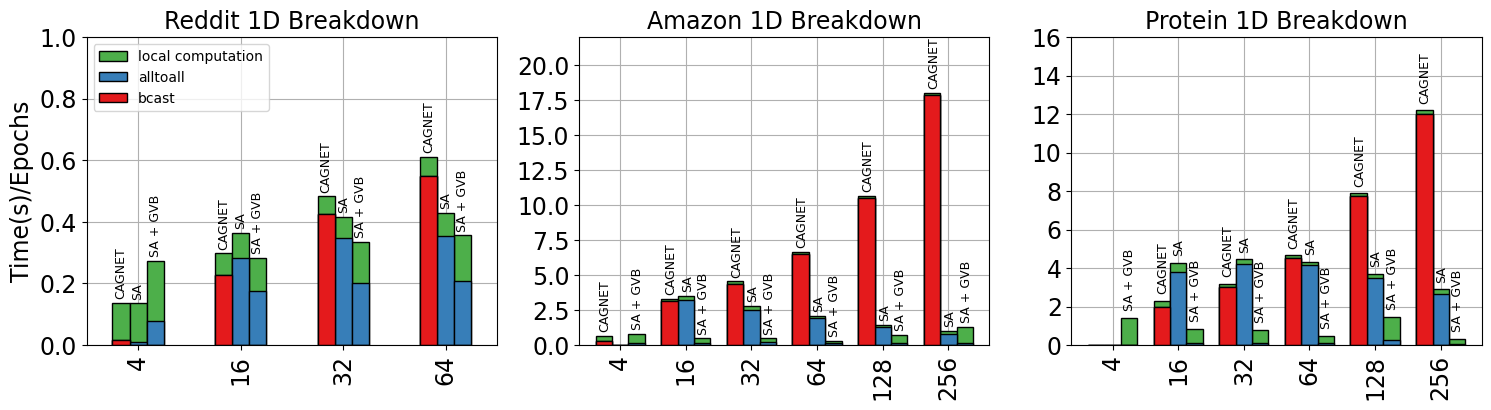}
    \caption{1D performance breakdown. The x-axis of each plot refers to the number of GPUs used. This breakdown includes \textit{local computation}, \textit{alltoall}, and \textit{bcast}. We compare results against CAGNET ~\cite{tripathy2020reducing}. SA represents just a sparsity-aware implementation, and SA + GVB refers to our sparsity-aware implementation used in conjunction with GVB graph partitioning. The sparsity-oblivious CAGNET implementation involves the broadcast and local computation, which in this case consists of the local SpMM computations. The sparsity-aware implementations used in the middle and right bar involves a single all-to-all call and a series of local computations, which includes gathering the data to send, allocating space in GPU memory, and the local SpMM computation.}
    \label{fig:1dbreakdown}
\end{figure*}

We compare the performance of our 1D sparsity-aware training against the 1D sparsity-oblivious training in Figure~\ref{fig:1dline}. 
We compare three schemes: the sparsity-oblivious training denoted as CAGNET, the sparsity-aware training described in Section~\ref{sec:spaware-gnn} and denoted as SA, and its enhancement with graph partitioning described in Section~\ref{sec:gp} and denoted as SA$+$GVB.
We plot the average time spent per epoch during the training with these schemes against the number of GPUs.

We can see that the original sparsity-oblivious gets slower as additional GPUs are used. The bandwidth costs do not scale with the number of GPUs, and the latency costs increase with $P$. The Reddit dataset, is small enough that training is latency-bound. Regardless of the algorithm or the number of processes, training time for one epoch takes less than a second. The Amazon dataset shows that for a small number of processors ($p=16$), the sparsity-aware algorithm makes little to no difference to the resulting training time. This means that the block rows are wide enough to the end that the number of nonzero subcolumns is not significantly less than the total number of subcolumns in that block. The benefit of sparsity-aware algorithms is clearer at higher process counts ($p \geq 32$) where communication is proportional to the edgecut. The same pattern appears in the Protein results where for lower process counts ($p < 64$), the sparsity-aware algorithm takes longer than the original sparsity-oblivious algorithm. In these cases, the cost of using point-to-point communication. which scales linearly based on the amount of data versus of broadcasts which scale logarithmically, is not displaced by communicating only rows that correspond to nonzero subcolumns. Like before, this is because the edgecut is not small enough. Interestingly, the sparsity-aware timing does not seem to be increasing either for lower process count. As $p$ increases ($p>64$), the sparsity-aware implementation starts to show benefit and the timing per epoch shows a promising decreasing trend. 

Figure ~\ref{fig:1dbreakdown} shows a granular timing breakdown. The original algorithm timings are overwhelmingly dominated by communication. While the Reddit data is latency-bound, at a higher process count ($p\geq 32$), the data shows cost of communication is decreasing. This is more prevalent in the Amazon dataset, where comparing just the original sparsity-oblivious implementation and sparsity-aware implementation, communication costs start to decrease at $p \geq 32$. The local computation cost stays about the same because it is mostly made of the local SpMM computation which is common across both implementations. 

\begin{figure}[!t]
    \includegraphics[scale=0.32]{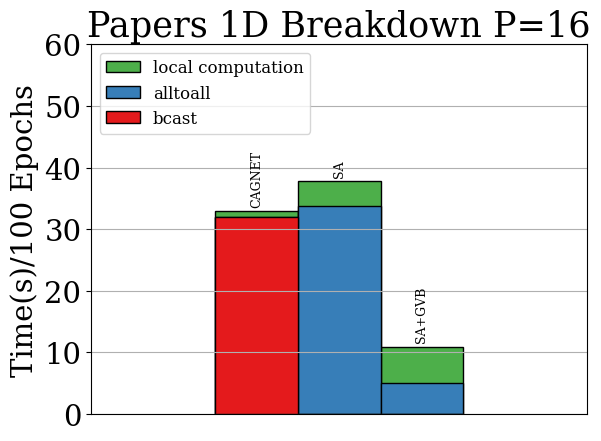}
    \caption{1D performance results for sparsity-oblivious, sparsity-aware, and graph partitioning for Papers dataset for $p=16$ processes. This breakdown includes \textit{local computation}, \textit{alltoall}, and \textit{bcast}, exactly like Figure~\ref{fig:1dbreakdown}.}
    \label{fig:1dbreakdownpapers}
\end{figure}

We also include results for our largest dataset, Papers, in Figure~\ref{fig:1dbreakdownpapers} on $p=16$ processes. We see that, like each of the other datasets, the sparsity-aware implementation with graph partitioning outperforms the sparsity-unaware version. In this case, there is a roughly $2.3\times$ improvement for similar reasons to other datasets (i.e., the \textit{alltoall} time is reduced). The caveat with graph partitioning is that it is a memory-intensive process. As such, running GVB ran out of memory when trying to partition Papers into more than $16$ partitions. 

\subsubsection{Graph Partitioning Performance}
We use the Graph-VB (GVB) graph partitioner~\cite{acer2016improving} with our 1D sparsity-aware GNN training to redistribute the sparse adjacency matrix.
As seen in Figure~\ref{fig:1dline}, using GVB greatly reduces training times.
The main reason for this can be seen Figure~\ref{fig:1dbreakdown}, where the communication bottleneck is largely overcome with the help of the partitioner.
Here, we can see that SA$+$GVB improves on SA by roughly $2\times$ across GPU counts for both Reddit and Amazon.
%
%
The degree of reduction in communication is dependent on the sparsity pattern of the graphs: Reddit and Amazon graphs are more irregular than the Protein graph, which makes the job of the partitioner easy in the latter and difficult in the former.
With Protein, the regularity improves the partitioning output enough that SA$+$GVB improves on SA by $14\times$ on $256$ GPUs.
However, GVB may sometimes increase the local computation time (compare the blue and green bars of SA and SA$+$GVB in Figure~\ref{fig:1dbreakdown}).
This is because of a rather loose constraint on computational load balance in partitioning in favor of further decrease in communication costs.
Since SA is oblivious to reducing communication while distributing the sparse matrix, it can solely focus on, and obtain better computational load balance than SA$+$GVB.

\begin{figure}[!t]
    \centering
    \includegraphics[scale=0.32]{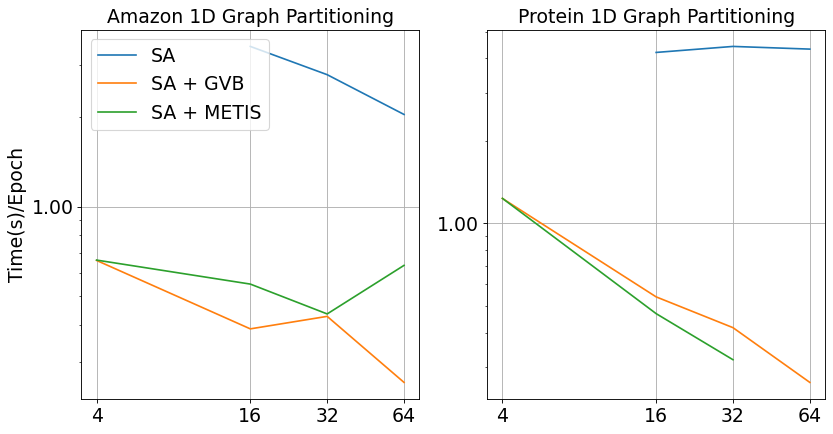}
    \caption{Graph partitioning performance results. The x-axis represents the number of GPUs used. The y-axis is the time per epoch. For this comparison, we use $p=4,16,32$, and $64$. Note that this is also a log-log plot. The missing line segments represent that this trial of the experiment ran out of memory. }
    \label{fig:1dgvbmetis}
\end{figure}

We next compare Graph-VB (SA$+$GVB) against METIS (SA$+$METIS) to assess the effect of addressing multiple cost metrics in reducing communication and present the results on Amazon and Protein in Figure~\ref{fig:1dgvbmetis}.
In the Amazon graph it is clearly seen SA$+$GVB results in lower training time by successfully reducing the maximum communication volume by a process, sometimes leading to more $2\times$ performance benefit on $64$ GPUs.
As noted by the METIS partitioning data in Table~\ref{tb:comm-imb-amazon-metis}, Amazon is a much more irregular graph.
The maximum communication volume by a process with $64$ GPUs is roughly $2\times$ larger than the average process's communication volume.
This imbalance makes Amazon training times speed up significantly with GVB over METIS.
%
%

With our Protein dataset, both partitioners exhibit similar behavior.
In this instance both partitioners reduce the edgecut drastically (only a few thousand edges become cut out of hundreds of millions edges), due to regularity in the Protein dataset.
Hence, the determining factor in training time between these two schemes becomes the computational load balance, in which SA$+$GVB performs slightly worse since it relies on variants of multi-constraint partitioning.
These results show that it is possible to further improve the training performance with a more capable partitioner than a plain partitioner, especially on difficult instances whose sparsity pattern is more irregular.

\subsection{Performance of 1.5D Sparsity-Aware Algorithm}

\begin{figure}
    \includegraphics[scale=0.32]{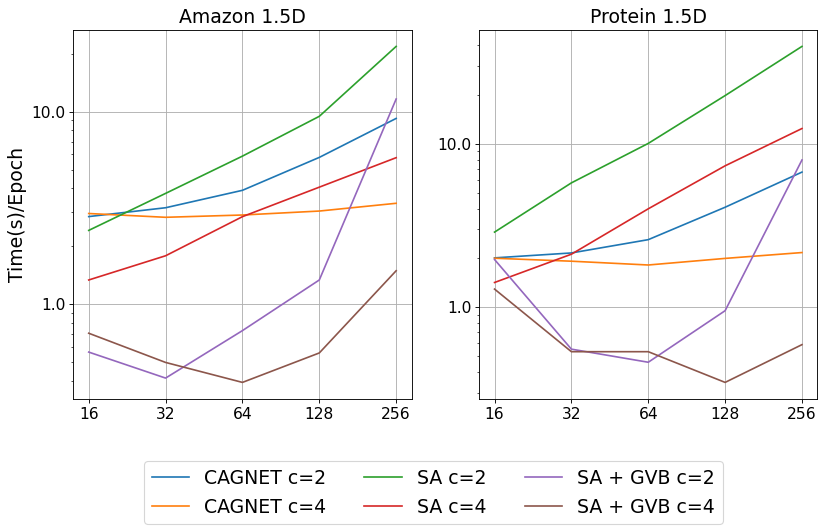}
    \caption{1.5D performance results for sparsity-oblivious, sparsity-aware, and graph partitioning for Amazon and Protein datasets for $c=2$ and $c=4$. The x-axis of each plot refers to the number of GPUs used. We use $p=16, 32, 64, 128$, and $256$. Note that $c$ represents the replication factor and that this is a log-log plot.}
    \label{fig:15dline}
\end{figure}

Figure~\ref{fig:15dline} shows training time results for the sparsity-oblivious, sparsity-aware, and sparsity-aware with GVB partitioning implementations on Amazon and Protein, with replication factors $c=2,4$. Note that the 1.5D with $c=1$ is identical to the 1D algorithm. For both the Amazon and Protein datasets, the sparsity-aware algorithm does not outperform the original sparsity-oblivious algorithm. We conclude that this is because the time taken by all-reduce in our 1.5D algorithm exceeds the time taken to send rows of $\mH$. In the sparsity-oblivious algorithm, increases $c$ decreases the broadcast time but increases the all-reduce time. However, the broadcast time is substantial, and increasing $c$ normally decreases the overall runtime. The sparsity-aware implementation reduces the time taken by the broadcast by only communicating necessary rows of $\mH$, to the point where the all-reduce call is expensive in comparison to sending necessary rows of $\mH$. This is observed clearly in the original algorithm as well as the sparsity-aware version. The sparsity-aware algorithm on the graph partitioned dataset reveals much better runtimes both for Amazon (a larger, but less dense graph) and Protein (a smaller, but far denser graph). Note that when using graph partitioning, we require $k=p/c$ partitions (rather than $p$ partitions as in the 1D algorithm) and the edgecut only decreases up to a certain point until it starts increasing again. This point depends on the input graph. Thus, we expect that the sparsity-aware algorithm combined with graph partitioning will have decreasing runtimes until $p=kc$ after which point, the runtime will start increasing again. We see this occur quite clearly with the Amazon dataset, where the minimum runtime for $c=2$ occurs at $p=32$ and for $c=4$, $p=64$. While this pattern is not as visible in the Protein data, we can see the formation of a minimum, signaling that there is an optimal number of partitions. 

\section{Conclusion}
\label{sec:conc}
In this work, we have shown a sparsity-aware approach to reducing communication between processors during GNN training that improves upon prior sparsity-oblivious work. We evaluated the sparsity-aware approach against four datasets (Reddit, Amazon, Protein, and Papers) of different size and density revealing that with 1D vertex partitioning, there is an overwhelming benefit in communication and training times. We also show that the graph partitioners that minimize the maximum communication volume are preferable to ones that minimize total communication volume (e.g. METIS, which is the most common partitioner used in GNN training). Using a graph partitioner that optimizes for both total communication volume (total number of edges crossing partitions) and maximum communication volume (maximum number of edges from one partition to another), we reduce load imbalance in communication and further reduce runtimes.
Our results with respect to the 1.5D algorithm show that the same idea of sparsity-awareness combined with graph partitioning can be applied to other communication-avoiding partitioning schemes, such as 2D, 2.5D, or 3D~\cite{Geijn1997,solomonik2013cyclops,azad2016exploiting}. All of our code is open-sourced at \url{https://github.com/PASSIONLab/CAGNET}.

\section{Acknowledgements}
This work is supported by the Advanced Scientific Computing Research (ASCR) Program of the Department of Energy Office of Science under contract No. DE-AC02-05CH11231. This research used resources of the National Energy Research Scientific Computing Center (NERSC), a Department of Energy Office of Science User Facility using NERSC award ASCR-ERCAP0033069.

\bibliography{sagnn}
\bibliographystyle{ACM-Reference-Format}

\end{document}